\documentclass[10pt,conference]{IEEEtran}
\IEEEoverridecommandlockouts
\usepackage{cite}
\usepackage{amsmath,amssymb,amsfonts,amsthm}
\usepackage{algorithm}
\usepackage{algorithmicx}
\usepackage{algpseudocode}

\usepackage{graphicx}
\usepackage{textcomp}
\usepackage{xcolor}
\usepackage{hyperref}
\hypersetup{hypertex=true}
\def\BibTeX{{\rm B\kern-.05em{\sc i\kern-.025em b}\kern-.08em
    T\kern-.1667em\lower.7ex\hbox{E}\kern-.125emX}}

\theoremstyle{definition}
    
\newtheorem{theorem}{Theorem}[section]
\newtheorem{definition}{Definition}
\newtheorem*{problem}{Problem}
\newtheorem*{myproof}{Proof}

\begin{document}

\newcommand{\ml}[1]{{\color{red}\bf [Meng: #1]}}
\newcommand{\red}[1]{{\color{red}\bf (#1)}}


\title{\textbf{Memory-aware Scheduling for Complex Wired Networks with Iterative Graph Optimization}}

\author{Shuzhang Zhong, Meng Li, Yun Liang, Runsheng Wang, Ru Huang}
\author{\IEEEauthorblockN{Shuzhang Zhong$^{1,2}$, Meng Li$^{2,1*}$, Yun Liang$^{1}$, Runsheng Wang$^{1,3,4}$, Ru Huang$^{1,3,4}$}
\IEEEauthorblockA{$^1$School of Integrated Circuits, Peking University, Beijing, China}
\IEEEauthorblockA{$^2$Institute for Artificial Intelligence, Peking University, Beijing, China}
\IEEEauthorblockA{$^3$Institute of Electronic Design Automation, Peking University, Wuxi, China}
\IEEEauthorblockA{$^4$Beijing Advanced Innovation Center for Integrated Circuits, Beijing, China}
\IEEEauthorblockA{meng.li@pku.edu.cn}
\thanks{* Meng Li is the corresponding author. This work was supported in part by the National Key R\&D Program of China (2020YFB2205502), and the 111 Project (B18001).}
}

\maketitle

\begin{abstract}

  Memory-aware network scheduling is becoming increasingly important for deep
  neural network (DNN) inference on resource-constrained devices. However,  
  due to the complex cell-level and network-level topologies, memory-aware
  scheduling becomes very challenging. While previous algorithms all suffer from
  poor scalability, in this paper, we propose an efficient memory-aware
  scheduling framework based on iterative computation graph optimization.
  Our framework features an iterative graph fusion algorithm that simplifies
  the computation graph while preserving the scheduling optimality.
  We further propose an integer linear programming formulation
  together with topology-aware variable pruning to schedule the simplified
  graph efficiently. We evaluate our method against prior-art algorithms on
  different networks and demonstrate that our method outperforms existing
  techniques in all the benchmarks, reducing the peak memory footprint by
  13.4\%, and achieving better scalability for networks with complex network-level topologies.
\end{abstract}

\section{Introduction}
\label{sec:intro}


Recent years have witnessed an increasing demand to deploy deep neural networks (DNNs) on resource-constrained edge devices with limited memory.
However, the topology of state-of-the-art (SOTA) DNNs is getting more and more complex.
On one hand, neural architecture search (NAS) is widely used to optimize the network's basic building blocks, i.e., cells,
and generates extensive irregular wirings within each cell, e.g., NASNet \cite{zoph2018learning}, DARTS \cite{liu2018darts}, Amoebanet \cite{real2019regularized}, etc;
while on the other hand, at the network level, multi-scale networks, e.g., HRNet \cite{wang2020deep}, NASFPN \cite{ghiasi2019fpn},
introduce dense connections among cells for more expressive feature extraction.

\begin{figure}[!tb]
    \centering
    \includegraphics[width=0.9\linewidth]{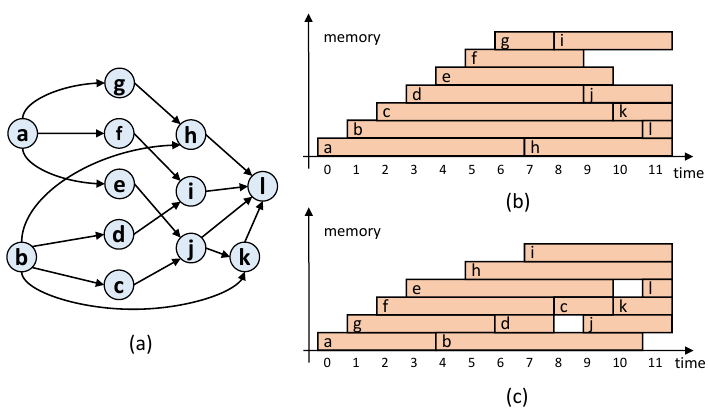}
    \caption{Scheduling directly impacts the peak memory footprint for complex
    networks.}
    \label{fig:cell_example}
\end{figure}

The complex DNN topology makes it challenging for the network inference to satisfy the memory constraints 
and calls for memory-aware optimizations \cite{ahn2020ordering,wang2022hierarchical}.
As shown in Figure~\ref{fig:cell_example}, the complex DNN topology forces to store more tensors
in the memory and hence, leads to a higher memory requirement.



Memory-aware scheduling determines the execution order for each operator in a DNN while minimizing the execution memory.
As in Figure~\ref{fig:cell_example} (b) and (c), given the same network, the two scheduling schemes both produce correct
results while the peak memory of schedule scheme in Figure~\ref{fig:cell_example} (c) is smaller.
Though promising, memory-aware scheduling for DNNs with complex topologies is very challenging.
First, NAS tends to generate complex cells with much more operators compared to hand-crafted cells, e.g., ResNet \cite{he2016deep} and VGG \cite{simonyan2014very}.
For example, NASNet has 40 operators per cell. In contrast, ResNet and VGG cells have 2 or 3 operators.
Meanwhile, while traditional networks leverage an almost linear topology both within each cell and across cells,
NASNet and HRNet introduce much more branches.
The large number of operators as well as irregular branches significantly increase the scale and complexity of the memory-aware scheduling problem.

Faced with these challenges, existing scheduling algorithms either suffer from a high execution memory or incur long scheduling time.
Deep learning compilers including Tensorflow \cite{abadi2016tensorflow} and TVM \cite{chen2018tvm} are unaware of execution memory footprint and use simple
topology-based scheduling strategy. SERENITY \cite{ahn2020ordering} is the first systematic memory-aware scheduling problem and proposes an approximate
dynamic programming (DP)-based algorithm to reduce the scheduling time. HMCOS \cite{wang2022hierarchical} further proposes a hierarchical approach that
simplifies the network computation graph through operator grouping and iteratively disassembles the graph for scheduling.
Though promising, both SERENITY and HMCOS rely on heuristics to simplify the computation graphs and suffer from a long scheduling time 
(more than 10 hours) when applied to multi-scale networks, e.g., HRNets.

In this paper, we examine the origin of the long scheduling time for SERENITY and HMCOS and observe 
that both methods heavily rely on
the simple, almost linear connections among cells to simplify the computation graph.
In contrast, the parallel branches of HRNet break the assumption and create a much larger and more flexible scheduling space.
To enable more efficient memory-aware scheduling, 
we propose an iterative pre-processing algorithm to optimize the complex DNN computation graph.
Our framework first simplifies the computation graph with iterative memory-aware fusion with provable optimality guarantee.
Our framework leverages an integer linear programming (ILP) formulation\cite{schrijver1998theory}  for scheduling and further proposes topology-aware variable pruning to speed up the solving process.
The paper makes the following contributions:
\begin{itemize}
    \item Memory-aware graph fusion. We propose iterative strategies to simplify the network computation graph while provably preserving the peak memory footprint.
    \item ILP-based Scheduling. We propose an ILP-based formulation for network scheduling and further improve the ILP efficiency with 
        topology-aware variable pruning.
    \item Memory-aware graph partition. We propose a memory-aware graph partition method for memory-constrained scheduling. The partition method is suitable for all hierarchical DNNs.
    \item Evaluation. We evaluate our framework on different complex DNNs and compared them to prior art methods, including topology-based algorithms, SERENITY, and HMCOS.
        Our framework achieves 13.4\% peak memory reduction compared to the topology-based algorithms and better scalability for networks with complex network-level topologies compared to SERENITY and HMCOS.
\end{itemize}

\section{Preliminaries}
This section introduces some basic concepts about memory-aware neural network scheduling.

\begin{definition}[Computation Graph]
A computation graph is a directed acyclic graph (DAG) $G=(V, E)$, in which each node $v \in V$ represents an operator and each directed edge $e \in E$ represents a precedence relation of the NN, in the form of input and output tensor.
Due to a tensor can be the input of multiple operators, it is a hyper-graph, in which an edge can connect any number of nodes.

A node $v$ is an ancestor of node $u$ if there is a directed path from the input to $u$ (excluding $u$) that includes $v$. 
The ancestor set of node $u$ is denoted as $anc(u)$. Similarly, $u$ is a descendant of $v$. The descendant set of node $v$ is denoted as $des(v)$.
If node $v$ is neither an ancestor nor a descendant of node $u$, it is considered a parallel node of $u$. The parallel node set of $u$ is denoted as $para(u)$.

A schedule of $G$ is a linear ordering of its nodes such that for every directed edge $(u, v) \in E$, vertex $u$ comes before vertex $v$ in the ordering.
It can be represented as a bijection $\sigma : V \leftrightarrow \{1,..., |V|\}$. This bijection assigns a unique number to each vertex in the DAG, indicating its position in the topological ordering.
\end{definition}

\begin{definition}[Memory Footprint]
A tensor x is stored in the memory between [$start(x), end(x)$], from the execution of its producer to the execution of its last consumer.
Given a computation graph and one of its topological scheduling, there are $2|V|+1$ memory usage: the memory usage states when executing an operator, and the memory usage states between executing two operators. We define \textbf{stable memory footprint} and \textbf{transient memory footprint} to describe the two states.
\begin{align}
    &s_{g, \sigma}^i = (\sum\limits_{start(x) \leq i \leq end(x)} size(x))+size(\sigma (i))\quad 1\leq i \leq |V| \nonumber \\
    &t_{g,\sigma}^i = \sum\limits_{start(x) \leq i \leq end(x)-1} size(x)\quad 0\leq i \leq |V| \nonumber
\end{align}\par
Given a computation graph $g$ and one of its topological sort $\sigma$, $s_{g, \sigma}^i$ means the stable memory footprint at the time $i$, including the live tensors and the extra memory usage when executing the operator. $t_{g, \sigma}^i$ means the transient memory footprint at time $i$, including the live tensors after executing the operator. Especially, $t_g^0$ means the input size of the graph.
\end{definition}

\begin{definition}[Induced Sub-graph]
Given a graph $G=(V, E)$ and its any subset of vertices $S \subset V$, the induced sub-graph $G_S = (S, E_S)$ is defined as the graph whose vertex set is $S$ and whose edge set is composed of all the edges whose endpoints are both in $S$, i.e. $E_S = \{\langle u, v\rangle|u, v\in S\, \langle u, v\rangle \in E\}$. The induced sub-graph is also a computation graph. 
\end{definition}

\begin{definition}[Isolated Sub-graph]
Given a graph $G=(V,E)$ and one of its induced sub-graph $G_U = (U, E_U)$. We define $G_U$ as an isolated sub-graph if it satisfies the following conditions:
\begin{align}
    &num(input(G_U))=1 \nonumber \\
    &num(output(G_U))=1 \nonumber \\
    &input(G_U).consumer \subset S \nonumber
\end{align}\par
Let $G_P$ be the induced sub-graph formed by the $P=para(U)$. Assume for the topological sort $\sigma$ of graph G, when executing the node $u \in U$, the previously executed node in P is p. Then the stable footprint in G is the sum of the stable footprint in $G_U$ and the transient footprint in $G_P$. Also, assume for the topological sort $\sigma$ of graph G, when executing the node $p \in P$, the previously executed node in U is u. Then the stable footprint in G is the sum of the stable footprint in $G_P$ and the transient footprint in $G_U$.
\begin{align}
    &s_{G, \sigma}^u = s_{G_U, \sigma}^u + t_{G_P, \sigma}^p \nonumber \\
    &s_{G, \sigma}^p = s_{G_P, \sigma}^p + t_{G_U, \sigma}^u \nonumber
\end{align}    
\end{definition}


\begin{problem}[Memory-Aware Network Scheduling]
Given a computation graph $G = (V, E)$, find a topological scheduling $\sigma$ that can minimize the maximum value of stable memory footprint.
\end{problem}

\section{Runtime Bottleneck for Network Scheduling}
\label{sec:motivating_exp}



The complexity of the memory-aware scheduling problem depends on both the scale and the architecture of the network.
While the impact of the network scale is easy to understand, in this section, we use a few simple network examples to
show how the network architecture impacts the scheduling complexity and how we can simplify the graph to speed up the scheduling process.

\begin{figure}[!tb]
    \centering
    \includegraphics[width=0.9\linewidth]{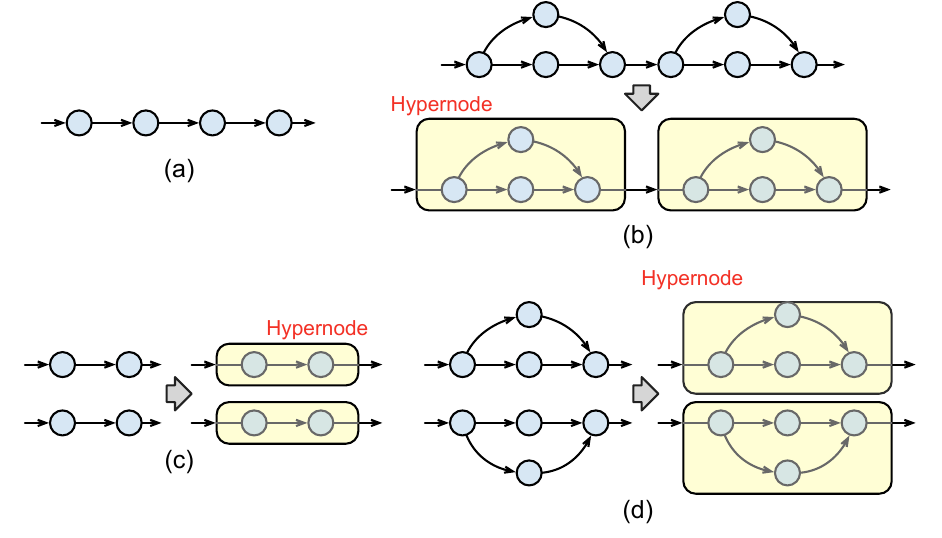}
    \caption{Examples of different network architectures: (a) linear network, (b) linear network with residual blocks,
         (c) 2-branch parallel network with residual blocks, and (d) 4-branch parallel network.}
    \label{fig:motivating_example}
\end{figure}

Network $A$ in Figure~\ref{fig:motivating_example}(a) has a simple linear network.
Considering the topological constraint, there is only 1 legal scheduling solution, which can be easily solved by topological sorting.
In Figure~\ref{fig:motivating_example}(b), Network $B$ adds residual blocks on top of Network $A$.
Because of the topological constraint, we can fuse each residual block into a hypernode and simplify the Network $B$ into a linear architecture.
For each hypernode, we can schedule its operators independently to get the transient and stable footprint 
and then, schedule the Network $B$ with hypernodes.
Note the fusion does not impact the optimality of the memory footprint for the final schedule regardless of the operator types or tensor sizes.

Network $C$ introduces 2 parallel branches in the architecture, while Network $D$ further adds residual blocks on top of Network $C$.
As there are fewer topological constraints in Network $C$ and $D$, their scheduling is much more flexible.
For example, Network $C$ has 6 scheduling candidates while Network $D$ has more than hundreds of candidates.
Hence, to get the scheduling with the minimum peak memory, more candidates have to be evaluated and compared,
leading to a significant increase in scheduling complexity.
If we can also fuse Network $C$ and $D$ to simplify their architectures, the scheduling candidate can be significantly reduced.
However, since there are no topological constraints, such simplification may lead to a larger peak memory.
Based on the example above, we have the following observations:
\begin{itemize}
    \item Network architecture directly impacts the complexity of memory-aware scheduling.
        Parallel branches have fewer topological constraints and higher scheduling freedom,
        which significantly increases the scheduling complexity.
    \item Sub-graph fusion can simplify the network architecture by merging a group of nodes into
        a hypernode. The main question, however, is how to select these nodes for fusion while
        preserving the optimality of the memory-aware scheduling.
\end{itemize}


\section{ILP-based Network Scheduling}
\begin{figure}[!tb]
    \centering
    \includegraphics[width=\linewidth]{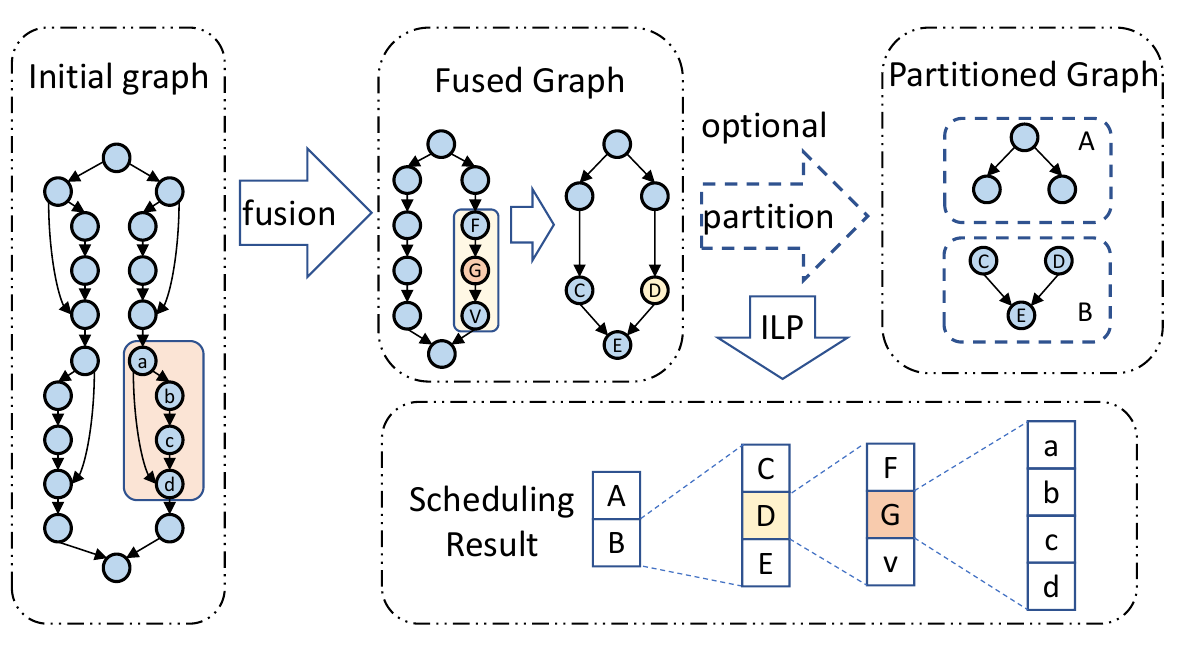}
    \caption{The brief flow of memory-aware scheduling. }
    \label{overview}
\end{figure}
The brief flow of our framework is shown in Figure \ref{overview}.
In this section, we first propose a topology-aware integer linear programming formulation for the memory-aware network scheduling problem. Then we propose the sub-graph fusion to simplify the graph with optimality guaranteed. In the end, we propose the memory-aware approximate graph partitioning to handle large-scale graphs.
\subsection{Topology-aware ILP Formulation}

\subsubsection{ILP Variable Definition}
The notations of ILP formulation are defined in Table \ref{tab:ilp_var}. Because one operator only has one output tensor, we denote the output tensor of operator $i$ as tensor $i$.
\begin{table}[!tb]
    \centering
    \caption{Notations used for the ILP formulation.}
    \begin{tabular}{cc}
    \hline 
        Variable & Meaning \\ \hline
        $O_{i, j}$ & operator $i$ is scheduled in step $j$ \\ 
        $T_{i, j}$ & tensor $i$ is stored in memory in step $j$ \\ 
        $S_{i}$ & size of tensor $i$ \\ 
        $ES_{i}$ & extra size of operator $i$ besides input and output \\ 
        $IN_{i}$ & set of input tensors of operator $i$ \\ 
        $INPUT$ & set of input tensors of graph \\ 
        $mem$ & peak memory footprint \\ \hline 
    \end{tabular}
    \label{tab:ilp_var}
\end{table}
\subsubsection{ILP Constraints}
This work aims to develop an efficient algorithm to determine the operator scheduling. It is necessary to ensure that the scheduling results satisfy the topology constraints of the computation graph. 

First, we need to guarantee the operators and execution times have a bijection relationship.
\begin{align}
    &\sum_{i} O_{i, j} = 1 \quad \forall j \in \{1,...|V|\} \label{ILP_1} \\
    & \sum_{j} O_{i, j} = 1 \quad \forall i \in V \label{ILP_2}
\end{align}\par
In the formulation, constraint \eqref{ILP_1} ensures in each step $j$, only one operator is scheduled; constraint \eqref{ILP_2} ensures each operator $i$ is scheduled only once.

Then we need to ensure the scheduling follows the topological order. 
\begin{align}
    & O_{i, j} \leq T_{k, j} \quad \forall k \in IN_{i} \label{ILP_3}\\
    & T_{i, j} \leq T_{i, j-1} + O_{i, j} \quad \forall i, j \label{ILP_4}\\
    & T_{i, 0} = 0 \qquad \forall i \notin INPUT \label{ILP_5}
\end{align}\par
In the formulation, constraint \eqref{ILP_3} indicates an operator $i$ can be scheduled in step $j$ only if all its input tensors are stored in memory in step $j$; constraint \eqref{ILP_4} indicates a tensor $i$ is stored in memory in step $j$ if it is already stored in step $j - 1$ or its producer operator $o_i$ is scheduled in step $j$; constraint \eqref{ILP_5} ensures each tensor other than input tensors cannot be stored in memory before the network is executed.

Based on the topological order constraints, we could express $mem$ as the peak memory footprint for the network inference.
\begin{align}
    &\sum_{i} (T_{i, j}\times S_{i}) + ES_{k}\times O_{k,j} \leq mem \quad \forall j,k
\end{align}

Finally, the objective is to minimize the peak memory footprint:
\begin{align}
    &\mathrm{min}  \quad mem \nonumber
\end{align}
\subsubsection{Variable Pruning}
\begin{figure}[!tb]
    \centering
    \includegraphics[width=\linewidth]{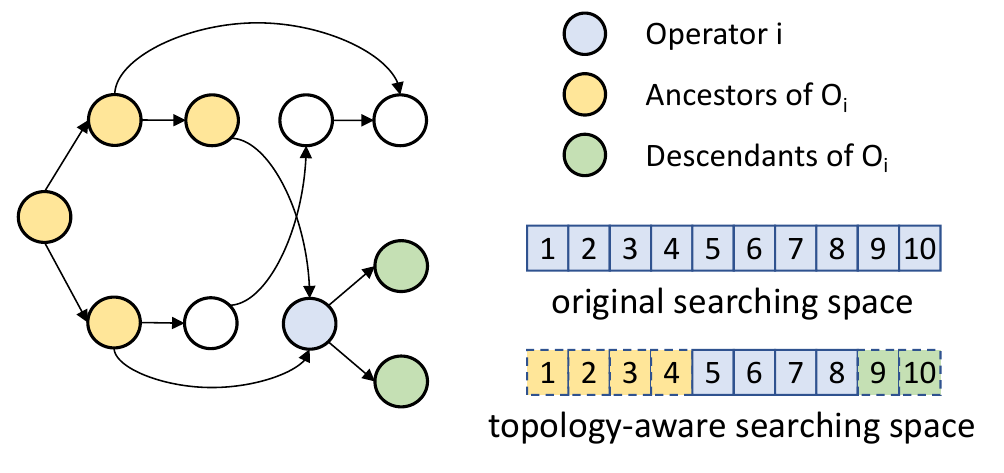}
    \caption{Topology-aware Variable Pruning. The searching space of $O_{ij}$ is limited between its ancestors and descendants.}
    \label{fig:range_limit}
\end{figure}

The variable complexity is $O(n^2)$ in the ILP formulation. However, the topology of the graph can provide more information for the integer programming formulations directly. For example, an operator can only be scheduled after the time step of its number of ancestors. As shown in \ref{fig:range_limit}, with 4 ancestors and 2 descendants, the operator $O_i$ can only be scheduled between time step $[5,8]$.

\par

Thus, we propose the topology-aware constraints to scale down the variable size.
\begin{align}
    &O_{i,j}=0\quad \forall \ i, 0< j \leq |anc(i)| \label{anc_op}\\
    &T_{i,j}=0\quad \forall \ i,0<  j \leq |anc(i)| \label{anc_tensor}\\
    &O_{i,j}=0\quad \forall \ i,|V|-|des(i)| <  j \leq |V| \label{des_op}\\
    &T_{i,j}=0\quad \forall \ i,|V|-max(des(use(i)))<  j \leq |V| \label{des_tensor}
\end{align}\par
Constraint \ref{anc_op}provides the prior knowledge for the ILP formulation that an operator must be executed after all of its ancestors have been executed. Constraint \ref{anc_tensor}provides the prior knowledge for the ILP formulation that a tensor cannot be stored in the memory unless all of its producer's ancestors have been executed. Constraint \ref{des_op} provides the prior knowledge for the ILP formulation that an operator must be executed before its descendants. Constraint \ref{des_tensor} provides the prior knowledge that a tensor can be removed after all of its child operators have been executed.

Variable pruning allows for the direct substitution of a significant number of variables with constants, effectively reducing the scale of the ILP problem that needs to be solved.

\subsection{Sub-graph Fusion with Optimality Guarantee}
\begin{figure}[!tb]
    \centering
    \includegraphics[width=0.9\linewidth]{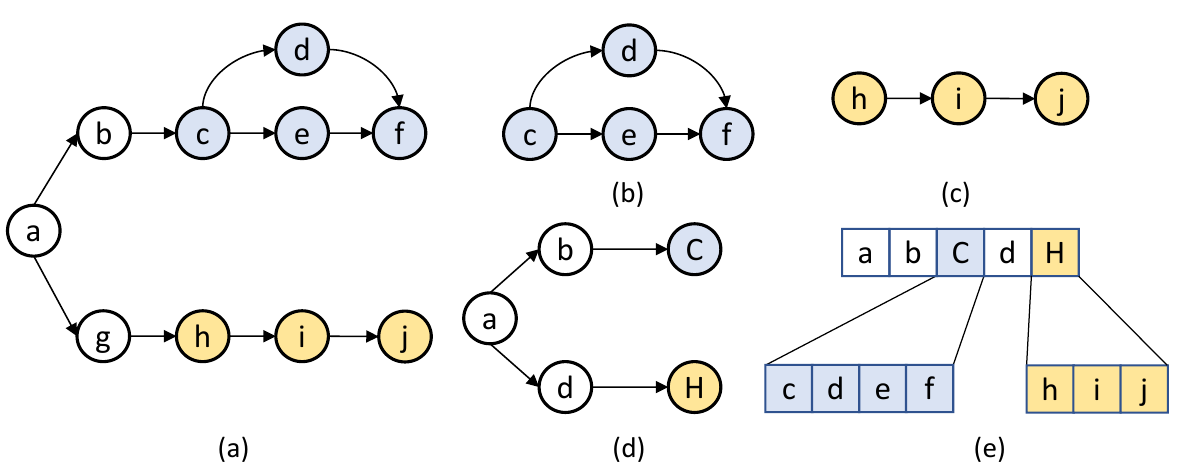}
    \caption{Example of sub-graph fusion: (a)original graph, (b)sub-graph C, (c)sub-graph H, (d)fused graph, (e)scheduling result.}
    \label{fig:fusion}
\end{figure}\par
The complexity of solving an ILP formulation increases exponentially with the scale of the graph. For this reason, we propose isolated sub-graph fusion to reduce the scale of the computation graph. 
The sub-graph fusion method replaces induced sub-graph into a node in the original computation graph. 
This will force the operators in the sub-graph to be executed one by one.

\begin{definition}[Fusion]
The fusion of isolated sub-graph $g=(U,E_g)$ in $G=(V,E)$ is to replace the sub-graph $g$ with a hyper-node $v_g$. The original graph $G$ will be replaced by the fused computation graph $G_f = (V_f, E_f)$, in which
\begin{align}
    &V_f =( V \setminus U) \cup \{v_g\} \nonumber \\
    &E_f = E \setminus \{<v_i,v_j>|v_i,v_j \in U\} \nonumber
\end{align}

Due to the fact that $v_g$ represents the entire sub-graph $g$, the stable memory footprint during the execution of $v_g$ is not simply the sum of the input and output tensors.
\begin{align}
    s_{G, \sigma}^{v_g} = &\sum\limits_{start(x) \leq \sigma(v_g) \leq end(x)} size(x) \nonumber\\
    &+mem_g-size(g.input)-size(g.output) \nonumber
\end{align}
\end{definition}

An example of sub-graph fusion is shown in Figure \ref{fig:fusion}. For the original computation graph shown in \ref{fig:fusion}(a), we can fuse the blue sub-graph C in \ref{fig:fusion}(b) and the yellow sub-graph H in \ref{fig:fusion}(c). The scheduling result of the induced sub-graph C is c-d-e-f, and the scheduling result of the induced sub-graph H is h-i-j.
Then the induced sub-graph C and H will be treated as a hyper-node respectively in \ref{fig:fusion}(d). Assume the scheduling result of (d) is a-b-C-d-H. Finally, replace the hyper-node with its inner scheduling result, and the scheduling result of \ref{fig:fusion}(a) is a-b-c-d-e-f-d-h-i-j. 

The sub-graph fusion introduces some manually specified scheduling order, which carries the risk of compromising the optimality of the scheduling results. Thus we first discuss the conditions that can keep the optimality when fusing linear sub-graph, then we extend it to the fusion of general induced sub-graph. 
\subsubsection{Monotonic Operator Fusion}
Given a computation graph $G = (V, E)$, a linear sub-graph is an inducing sub-graph $G_s = (S,E_S)$, 
with each edge $\langle s_i,s_{i+1} \rangle \in E$ and each operator only having one parent operator and one child operator.

A linear sequence of operators is ubiquitous in NNs, such as the basic conv+bn+relu operation. A large number of linear sequences of operators appear in the networks with the basic sequences, such as NASNet\cite{zoph2018learning}, HRNet\cite{wang2020deep}, etc.

\begin{theorem}
Given $G=(V,E)$ and a linear sequence $[v_1,v_2,...v_n]$ in $G$. The induced sub-graph of the sequence is $g=(V_S, E_S)$. The sequence can be fused if both of the following conditions are satisfied:
\begin{align}
    &t_{g, \sigma}^i\leq t_{g, \sigma}^j\quad 0\leq i <j\leq n\\
    &s_{g, \sigma}^i\leq s_{g, \sigma}^n\quad 0\leq i <j\leq n
\end{align}\par or
\begin{align}
    &t_{g, \sigma}^i\geq t_{g, \sigma}^j\quad 0\leq i <j\leq n\\
    &s_{g, \sigma}^0\geq s_{g, \sigma}^i\quad 0\leq i <j\leq n
\end{align}\par
\end{theorem}
In other words, the transient memory footprint of the sequence is monotonic. For the monotonic increase linear sequence, the costs of the previous operators are lower than the last operator, so the memory footprint of the fusion node can be regarded as the footprint of the last node. 

\subsubsection{Operator Fusion beyond Linear Structures}
Beyond the linear sequence operators, there are more general sub-graph structures in the real-world computation graph. We find out and prove the conditions of fusing general induced-graphs optimality.

\begin{theorem}
\label{theorem:general}
Given $G=(V,E)$ and one of its \textbf{isolated induced sub-graph} $g = (U,E_u)$. The sub-graph $g$ can be fused without compromising optimality if the following conditions are satisfied:
\begin{align}
    &t_{g, \sigma}^i \geq t_{g, \sigma}^0\quad 1\leq i < n \label{graph-input}   \\
    &t_{g, \sigma}^i \geq t_{g, \sigma}^n\quad 1\leq i < n \label{graph-output}
\end{align}
\end{theorem}
In another word, the middle of the transient footprint in the sub-graph is higher than the input transient footprint and output transient footprint. Therefore it conveys no benefits to stop in the middle, the sub-graph can be fused and executed at one time.

\begin{myproof}
Assume that the ancestor set of g is X, the descendant set of g is Z, and the parallel set of g is Y. The optimal schedule when fusing the sub-graph is:
\begin{align}
    &x_1...x_ny_1...y_ig_1...g_i...g_n...y_j...y_n...z_n\label{sub-graph_optimal}
\end{align}
\begin{itemize}
    \item If the peak memory footprint is at the execution of $g_i$ in the induced sub-graph g. The peak memory footprint is $s_{g,\sigma}^{g_i}+t_{y,\sigma}^{y_i}$, including the stable footprint in g and transient footprint in y. If breaking the fusion and moving the $g_i$ after $y_j$ can reduce the stable memory footprint when executing $g_i$ from $s_g^i+t_y^{y_i}$ to $s_g^i+t_y^{y_j}$. The new schedule is
\begin{align}
    &x_1...x_ny_1...y_ig_1...g_{i-1}...y_jg_i...g_n...y_n...z_n \label{sub-graph_not_optimal}
\end{align}    
However, according to the assumption, the peak memory of $...y_1...y_i...y_jg_1...g_n...$ is higher than \ref{sub-graph_optimal}. Therefore, there exists a $y_k$ between $y_i$ and $y_j$, whose stable footprint is higher than the peak memory footprint in \ref{sub-graph_optimal}. It is the sum of its stable footprint in y and the transient footprint in g, $s_y^{y_k}+t_g^0>s_g^{g_i}+t_y^{y_i}$. According to the condition, the memory in the middle of the sub-graph is higher than the input and output, i.e. $t_g^i>t_g^0$. Thus the stable footprint in \ref{sub-graph_not_optimal} of $y_k$ is $s_G^{y_k}=s_y^{y_k}+t_g^{i-1}>s_y^{y_k}+t_g^0>s_g^{g_i}+t_y^{y_i}$.
    \item If the peak memory footprint is at the execution of $y_j$ in the induced sub-graph Y and $\sigma (y_k) < \sigma (g_1)$, i.e. $s_{G,\sigma}^{y_j} = s_{y,\sigma}^{y_j}+t_{g,\sigma}^{g_0}$. If we break the execution of the sub-graph, the stable footprint in the execution of $y_j$ is $s_{G,\beta}^{y_j} = s_{y,\sigma}^{y_j}+t_{g,\sigma}^{g_i}$. However, According to \ref{graph-input}, $s_{G,\beta}^{y_j} = s_{y,\sigma}^{y_j}+t_g^0 \leq s_{y,\sigma}^{y_j}+t_{g,\sigma}^{g_i}$. So the scheduling sequence \ref{sub-graph_optimal} is the optimal scheduling result.
    \item If the peak memory footprint is at the execution of $y_j$ in the induced sub-graph Y and $\sigma (y_k) > \sigma (g_n)$. The proof is similar with when $\sigma (y_k) < \sigma (g_0)$.
    \item If the peak memory footprint of G is in the sub-graph X or Z. Because the operators in the induced sub-graph x are always scheduled before the execution of g, it has nothing to do with whether fuse the sub-graph g or not.  For the same reason, whether breaking the fusion of g or not does not influence the footprint of z. 
\end{itemize}
\end{myproof}
\begin{algorithm}
    \caption{Sub-Graph Fusion}
    \label{algorithm:fusion}
    \begin{algorithmic}
        \Require Original Graph G=(V,E), Max Sub-Graph Size M 
        \Ensure  Fused Graph G'=(V',E')
        \State $G'\gets G$, $G'' \gets None$
        \While{$G' \ne G''$}
        \State $G'' \gets G'$
        \For{$tensor$ in $G.tensors$}
            \State $fused\_ops \gets$ []
            \State $queue \gets$ tensor.consumer\_ops
            \While{$queue$.size $>$ 0}
                \State $op \gets$ bfs\_get(queue)
                \State $fused\_ops$.append(op)
                \If{$queue$.size==0 and $fused\_ops$.size$>$1}
                    \If{$fused\_ops$.legal}
                    \State $G'$.fuse($fused\_ops$)
                    \EndIf
                    \State \textbf{continue}
                \ElsIf{$fused\_ops$.size$>$M}
                \State \textbf{continue}
                \EndIf
                \State $queue \gets $ bfs\_add($queue$)
            \EndWhile
        \EndFor
        \EndWhile
        \State \Return $G'$
    \end{algorithmic}
\end{algorithm}

Algorithm~\ref{algorithm:fusion} shows the general induced sub-graph fusion algorithm. It fuse the sub-graphs iteratively until no additional legal sub-graphs remain. It enumerates each tensor as the input to try to find legal induced sub-graphs. In order to avoid the non-convergence of the search, it sets M as the maximum size of the sub-graph. If the number of  visited operators is higher than M, the search will be broken. 

\begin{figure}[!tb]
    \centering
    \includegraphics[width=\linewidth]{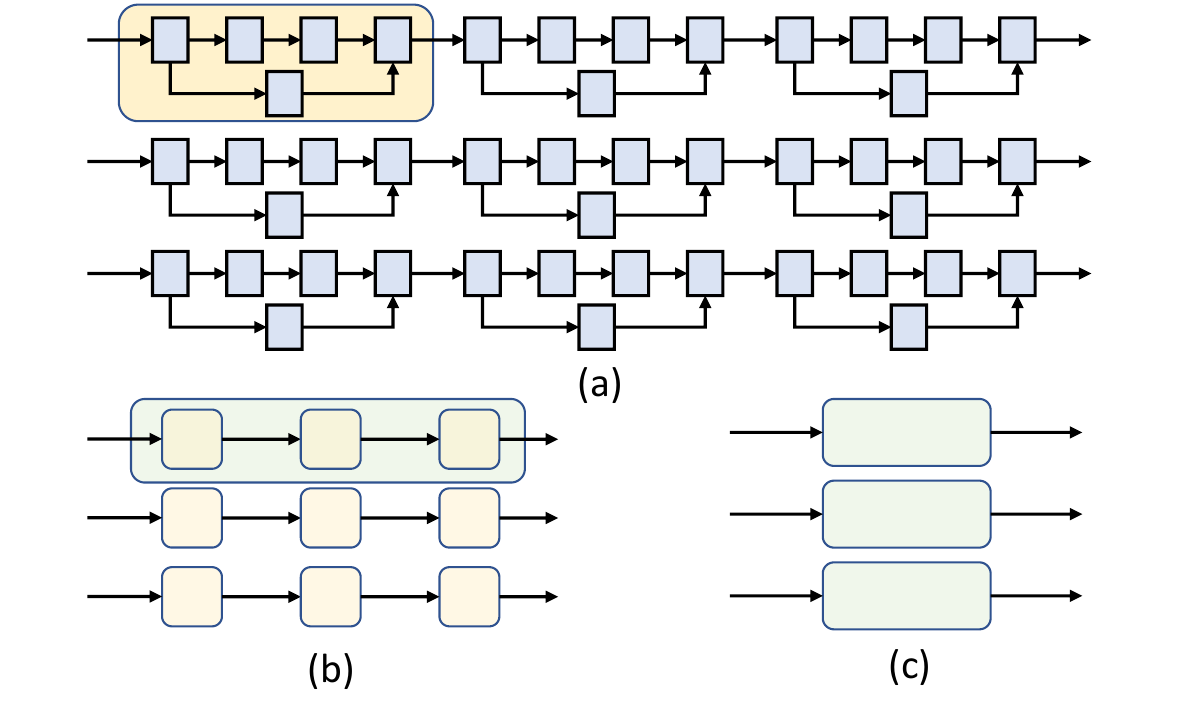}
    \caption{Iterative Fusion: (a) original high-resolution block, (b) high-resolution block after 1 cycle general operator fusion, (c) high-resolution block after 2 cycles general operator fusion.}
    \label{fig:sub-graph}
\end{figure}\par

Figure~\ref{fig:sub-graph} shows the iteration of the general operator fusion. The original high-resolution block as Figure~\ref{fig:sub-graph}(a) shows has three parallel linear residual blocks. The first cycle of fusion fuses the residual blocks as to Figure~\ref{fig:sub-graph}(b). The next cycle of fusion fuses the linear sequence of the hyper-nodes. The high-resolution block can be fused to only 3 operators with optimality guaranteed.   

\subsection{Memory-aware Graph Partitioning}

\begin{figure}[!tb]
    \centering
    \includegraphics[width=0.9\linewidth]{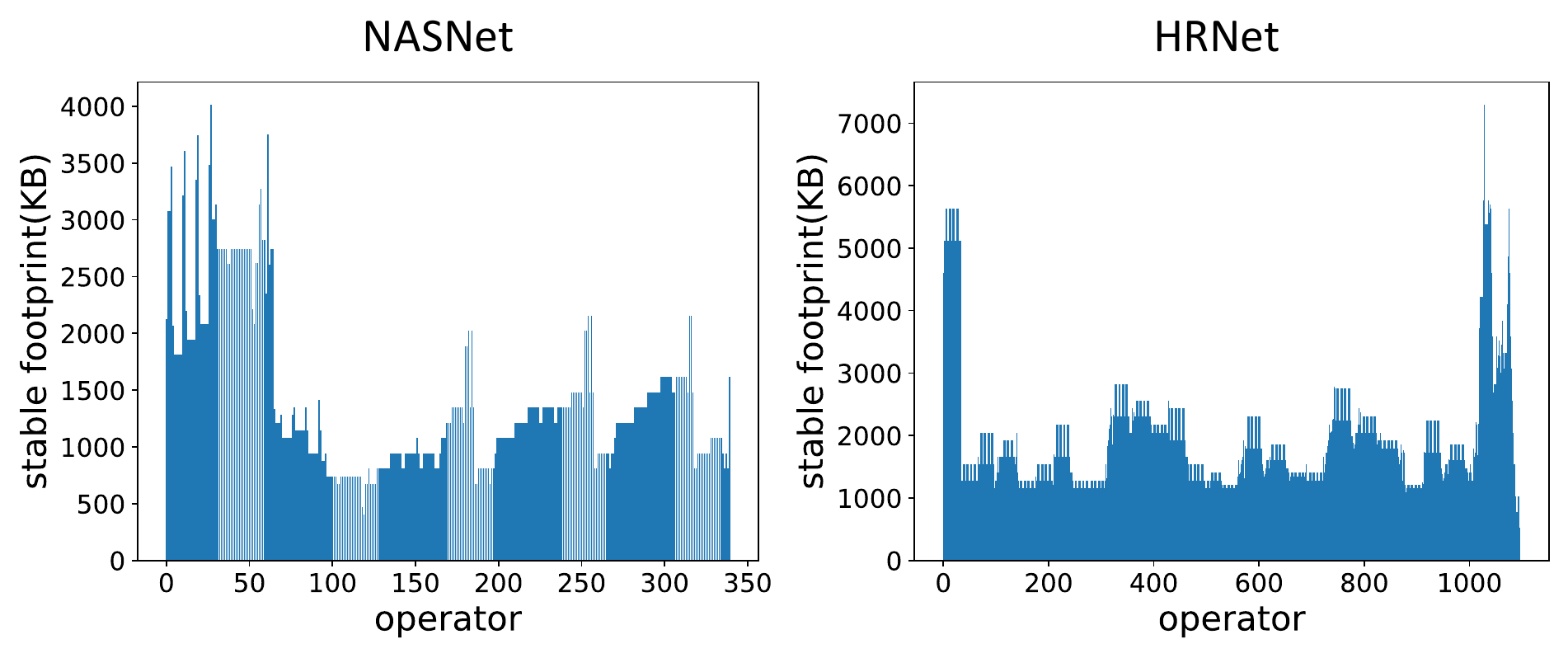}
    \caption{Stable memory footprint of NASNet and HRNet.}
    \label{fig:exp footprint}
\end{figure}

While the sub-graph fusion method can reduce the scale of the computation graph to a large extent, the ILP formulation may still suffer from the exponential complexity for NNs with a large computation graph.
We observe the peak memory usage of a DNN may occur only in a specific region of the graph. 
Hence, by focusing on optimizing the scheduling strategy for that particular region, it is possible 
to achieve an memory-efficient scheduling scheme without solving the entire computational graph. 
As shown in Figure~\ref{fig:exp footprint}, the operators closer to the input tend to have the highest stable
footprint in NASNet, and the operators closer to the output tend to have the highest stable footprint in HRNet.

Therefore, we propose a memory-aware acyclic graph partitioning algorithm.
Theorem~\ref{theorem:general} reveals that isolated sub-graphs with smaller input and output tensors tend to be fused as a whole, for it conveys no benefits to stop the execution of the sub-graph in the middle.
While the isolated sub-graph imposes a strong constraint on the fusion,
we now relax this condition but follow the same intuition to design a heuristic graph partitioning algorithm.
Let the size of the tensors be the edge weight, the memory-aware partitioning problem can be transferred to the min-cut partitioning problem.
Given a graph $G=(V, E)$, the $k$-way acyclic graph partitioning is a partition $\{V_1,V_2,...V_k\}$, in which for all $u,u'\in V_i$ and $v,v' \in V_j$
\begin{align}
    (u \in anc(v)) \wedge (u' \in des(v')) = \varnothing \quad \forall i,j\ i\ne j  \nonumber
\end{align}    
The condition ensures that there is no cross-dependency between any sub-graphs.

\begin{figure}[!tb]
    \centering
    \includegraphics[width=0.9\linewidth]{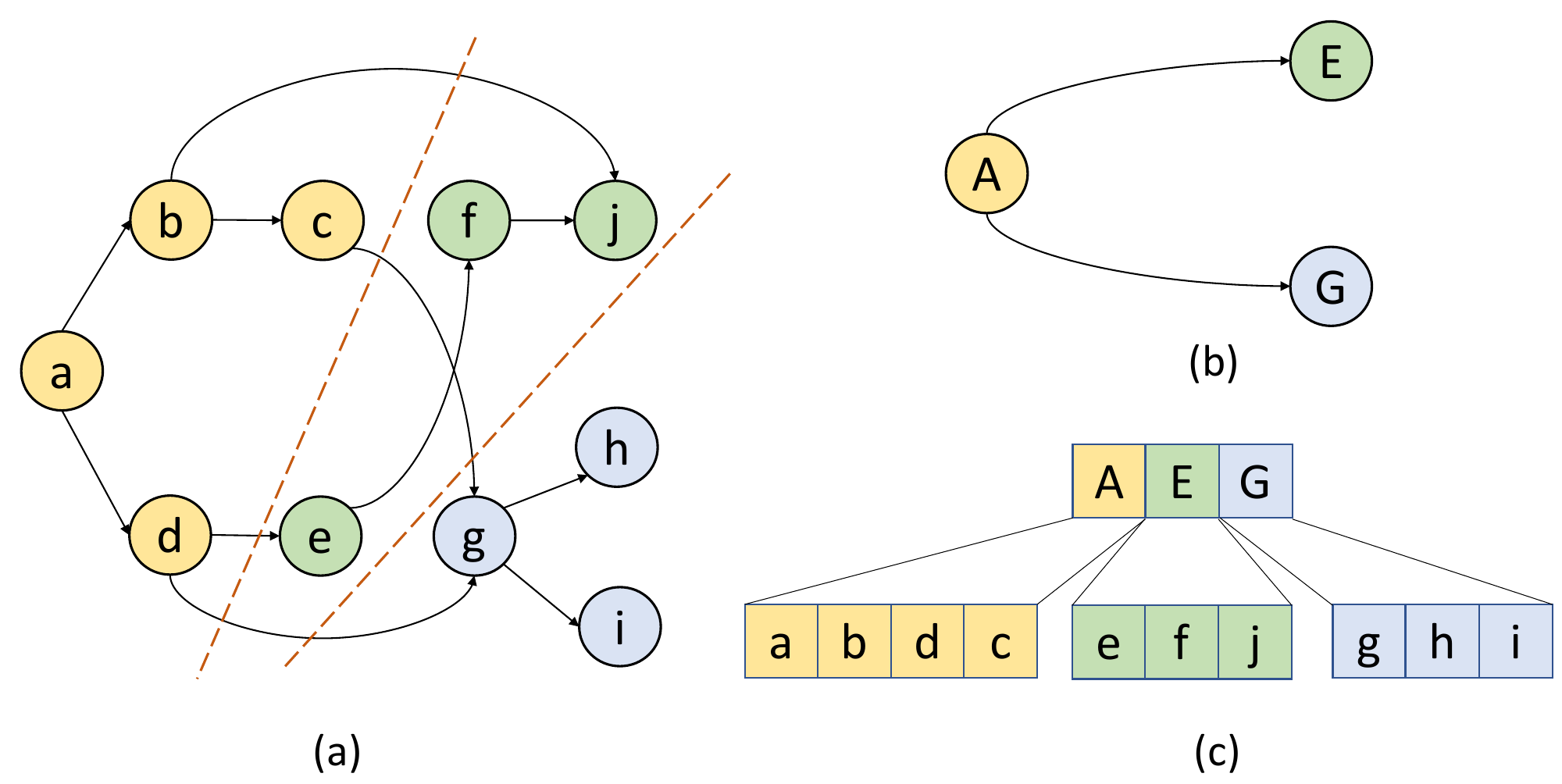}
    \caption{Example of graph partition: (a) original graph, (b) partitioned graph, (c) scheduling result.}
    \label{fig:partition}
\end{figure}

An example of graph partition is shown in Figure \ref{fig:partition}. The original NN computational graph (a) is partitioned into 3 sub-graphs. Each graph is scheduled seperately. The final result is the combination of each sub-graph.

The peak memory footprint tends to appear only in part of a computation graph. Therefore, we adopt the reverse post-order scheduling first to discover the potential operators that may affect the peak memory footprint. In the scheduling of the partitioned graph, we only use the ILP scheduling for the sub-graphs that contain these potential operators. For other sub-graphs that do not contain the operators, we adopt reverse post-order scheduling for them.

\begin{algorithm}
    \caption{Partitioning-based Scheduling}
    \label{algorithm:scheduling}
    \begin{algorithmic}
        \Require Origin Graph $G=(V,E)$, Number of Parts $k$ 
        \Ensure  Scheduling Result $s$
        \State $prev\_num \gets \infty$
        \While{len($G.ops$)$<prev\_num$}
            \State $prev\_num \gets$ len($G.ops$)
            \State Fuse the operators of G.
        \EndWhile
        \State $rpo\_schedule \gets $RPOSchedule($G'$)
        \State $peak\_op \gets $peak($rpo\_schedule$)
        \State $H \gets $hypergraph($G$)
        \State $\{V_1,...,V_k\} \gets $partition($H,k$)
        \State $s \gets $[ ]
        \For{$V_i$ in $\{V_1,...,V_k\}$}
            \State $G_i \gets G$.induce($V_i$)
            \If{$peak\_op$ in $V_i$}
                \State $sub\_schedule \gets $ILPSchedule($G_i$)
            \ElsIf{$peak\_op$ not in $V_i$}
                \State $sub\_schedule \gets $RPOSchedule($G_i$)
            \EndIf
            \State $s\gets s+sub\_schedule$            
        \EndFor
        \State \Return $s$
    \end{algorithmic}
\end{algorithm}

Algorithm~\ref{algorithm:scheduling} shows the complete scheduling algorithm. First, we use iterative operator fusion to lower the scale of the operators in the network. Then, we use the RPO schedule to discover the potential peak memory operators. Then, we convert the graph to a hyper-graph and partition it. Last, we schedule these sub-graphs by ILP or RPO according to the initial RPO scheduling result.

\section{Experimental Results}

\subsection{Experiment Setup}
We implement the ILP-based network scheduling algorithm with
gurobi\cite{bixby2007gurobi} and the acyclic partition by
dagP\cite{herrmann2017acyclic}. In this experiment, we compare our method
against topology-based methods and existing memory-aware scheduling methods.

\subsubsection{Benchmarks} We benchmark our ILP-based scheduler on three
classes of benchmarks, including NASNet,
Randwire and HRNet. NASNet and RandWire have similar characteristics with
complex topologies within cell and simple network-level connections \cite{xie2019exploring}.
In contrast, HRNet leverages simple ResNet blocks but the connections among
blocks are much more complex, including parallel branches, fusion branches,
etc.

\subsubsection{Baselines} We evaluate our method against reverse post order
(RPO), SERENITY, and HMCOS on the 3 classes of benchmarks above. RPO is a basic
topological sorting method which is simple but not memory-aware and gets wide
adoption in deep learning compilers such as Tensorflow-Lite\cite{abadi2016tensorflow}.
HMCOS \cite{wang2022hierarchical}, which is the
start-of-art memory-aware scheduling method constructs a hierarchical
computation graph and iteratively schedules the operators based on the dynamic
programming scheduling method proposed by SERENITY \cite{ahn2020ordering}.


\subsubsection{Metrics} We evaluate our method mainly with the peak memory
footprint and the scheduling time. 

\subsection{Main Results}

\subsubsection{Peak Memory Footprint Comparison}
Table~\ref{table:mem_footprint_comp} shows the peak memory footprints of
RPO, SERENITY, HMCOS, and our method. We also show the peak memory reduction of our
method against RPO. We set the maximum running time to 30 seconds. The current best 
result will be taken as the final result if exceeding the time limit.
The results show that our method can reduce the peak memory footprint by up to
26.4\% compared with TensorFlow Lite. Note that the peak memory footprint in
Darts is at the start of the network, thus the topological scheduling and
memory-aware scheduling have the same result. Our method can achieve the minimum footprint
in all benchmarks.

\begin{table}[!ht]
    \centering
    \caption{Peak memory footprint comparison with prior art methods.}
    \begin{tabular}{c|ccccc} 
    \hline \hline
    ~            & RPO   & SERENITY & HMCOS & Ours          & Reduction  \\
    \hline 
    NASNet-A     & 4405  & 4405     & 3943  & \textbf{3943} & 10.5\%        \\
    AmoebaNet-A  & 10510 & 10510    & 7732  & \textbf{7732} & 26.4\%        \\
    Darts        & 1764  & 1764     & 1764  & \textbf{1764} & 0\%           \\
    Randwire1    & 5494  & 5494     & 4538  & \textbf{4538} & 17.4\%        \\
    Randwire2    & 6927  & 6927     & 6210  & \textbf{6210} & 10.3\%        \\
    Randwire3    & 5255  & 5255     & 4777  & \textbf{4777} & 9.1\%         \\
    HRNet-W18-V1 & 7936  & 7936     & 7936  & \textbf{6368} & 19.8\%        \\
    HRNet-W18-V2 & 7936  & 7936     & 7752  & \textbf{6432} & 19.0\%        \\
    HRNet-W32    & 7936  & 7936     & 7808  & \textbf{7296} & 8.1\%         \\
    \hline
    Average      & ~     & ~        & ~     & \textbf{~}    & 13.4\%        \\
    \hline \hline
    \end{tabular}
    \label{table:mem_footprint_comp}
\end{table}

\subsubsection{Scheduling Time Comparision}
Table~\ref{table:scheduling_time} shows the scheduling time of HMCOS and Ours. 
As HMCOS takes special manual optimizations for NASNet, our
method has similar results compared with it.
For HRNet, our method can finish the scheduling
in 10 seconds, but HMCOS needs more than 30000 seconds to finish the scheduling.  
\begin{table}[!ht]
\caption{Scheduling time comparison with prior art methods.}
\centering
\label{table:scheduling_time}
\begin{tabular}{c|ccc}
\hline \hline
                   & HMCOS(s) & OURS(s) & Speedup     \\
\hline
NASNet-A           &1.51       &4.62      &0.33x             \\
AmoebaNet-A        &0.80       &0.11      &7.27x             \\
Darts              &0.18       &0.09      &2x             \\
Randwire1           &73       &23      & 3.17x\\
Randwire2        &25.04       &17.09      & 1.46x          \\
Randwire3              &1.39       &1.22      &1.13x           \\
HRNet-W18-Small-V1 &30000+   &0.96    &/             \\
HRNet-W18-Small-V2 &30000+        &5.01      & /            \\
HRNet-W32          &30000+       &3.20      & /            \\
\hline \hline
\end{tabular}
\end{table}

\subsubsection{Graph Scale Comparision}
\begin{figure}[!tb]
    \centering
    \includegraphics[width=0.9\linewidth]{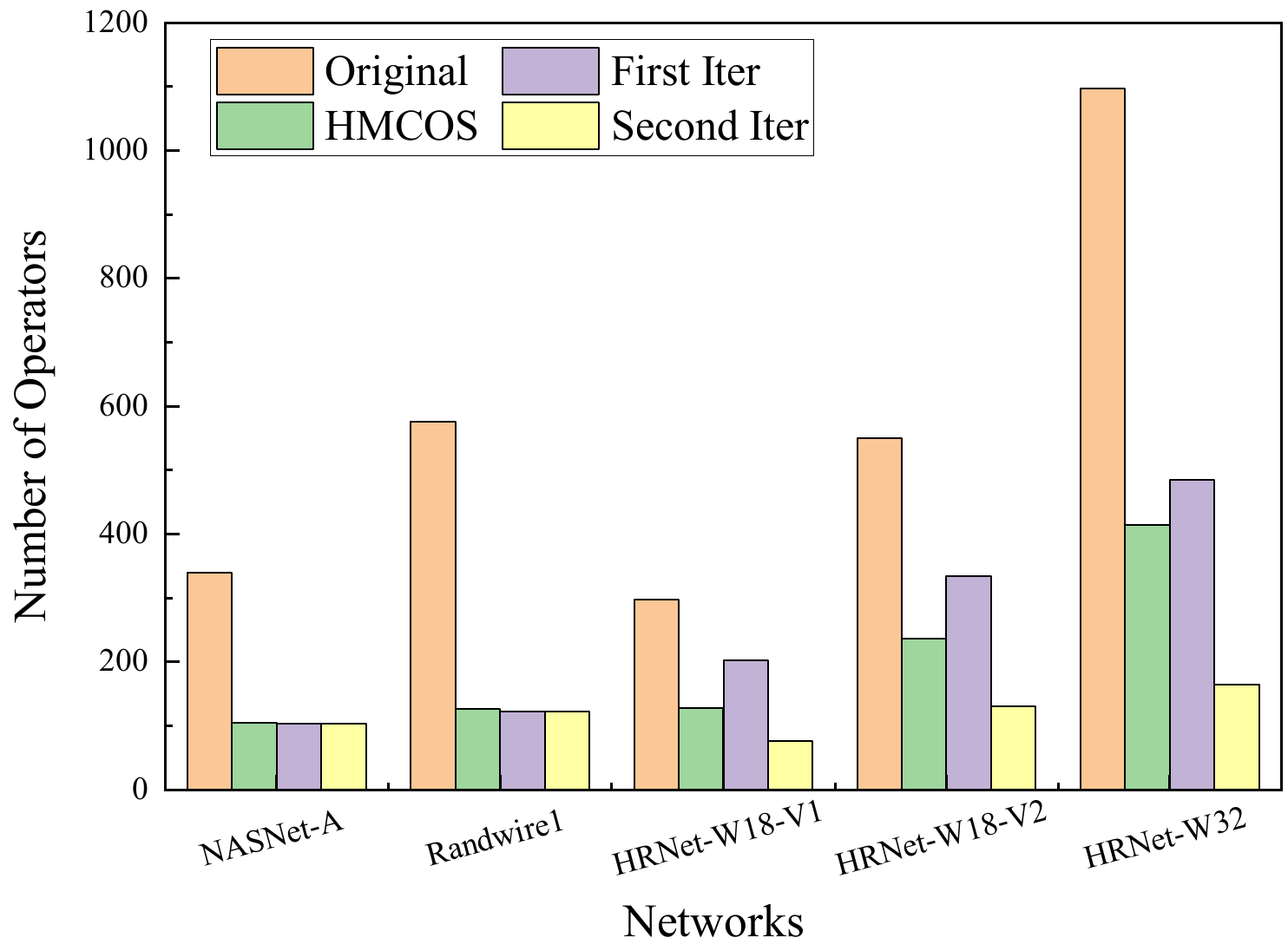}
    \caption{Graph Scale Comparison with Original Graph and HMCOS.}
    \label{fig:scale}
\end{figure}

To evaluate the effectiveness of our iterative graph optimization, 
we compare our optimized computation graph with the original graph and the one
optimized by HMCOS. 
This analysis aimed to provide further
insights into the impact and efficiency of our iterative graph optimization
approach. Figure~\ref{fig:scale} illustrates the operator count comparison.
As can be observed, our proposed iterative graph optimization method 
achieves a maximum reduction in scale
of up to 6$\times$ compared to the original computation graph and 4$\times$ 
compared to HMCOS while preserving the optimality of memory-aware scheduling.


\begin{table*}
\centering
\caption{Variable numbers in ILP formulation(marked as V), Time (in seconds, marked as T) for scheduling architectures and the reduction in Variable numbers and speedup in time against single partition method.}
\label{table:ablation}
\resizebox{\linewidth}{!}{
\begin{tabular}{c|c|cc|cccc|cccc|cccc} 
\hline \hline
Method & Origin & \multicolumn{2}{c|}{Partition} & \multicolumn{4}{c|}{Partition+Fusion} & \multicolumn{4}{c|}{Partition+Pruning} & \multicolumn{4}{c}{All}           \\ 
\hline 
       & T      & V    & T                       & V   & Reduction & T    & Speedup      & V    & Reduction & T   & Speedup          & V   & Reduction & T    & Speedup  \\
NASNet   & $>$3600   & 2903  & 213                    & 402  & 86.10\%   & 4.84 & 44.0x       & 2837 & 2.20\%    & 210  & 1.01x           & 590  & 79.70\%   & 1.47 & 144.5x   \\
Randwire & $>$3600   & 21120 & $>$3600                   & 2292 & 89.10\%   & 16   & /           & 9527 & 54.80\%   & $>$3600 & /               & 1809 & 91.40\%   & 1.75 & /        \\
HRNet    & $>$3600   & 2511  & 10.7                   & 310  & 87.60\%   & 2.7  & 3.96x       & 758  & 69.80\%   & 3.9  & 2.74x           & 277  & 88.90\%   & 0.96 & 11.1x    \\
\hline \hline
\end{tabular}
}
\end{table*}

\subsubsection{Ablation Study}
We further explore how the components of our method contribute to the result.
We use NASNet-A, Randwire1 and HRNet-W18-Small-V1 as an example and show the impact of the
proposed speedup techniques on 1) ILP problem size, and 2) scheduling time.

Table~\ref{table:ablation} shows the reduction of ILP problem size and the
speedup of scheduling time under variable pruning and iterative operator
fusion on the benchmarks. The optimality-guaranteed method can
scale down the problem size by $8 \sim 10 \times$ and achieve more than
$10\times$ speedup. 

We further explored the effectiveness of heuristic
memory-aware graph partitioning methods compared to the naive topology-aware graph
partitioning approach. The naive partitioning method involves partitioning the
graph solely based on its topology information. Table \ref{table:partition
method} shows that our heuristic memory-aware algorithm can achieve the optimal scheduling
result in most computation graphs. Our heuristic partitioning method
outperformed the naive method in all tested cases.

\begin{table}[!tb]
\centering
\caption{Comparison between heuristic and naive algorithm.}
\label{table:partition method}
\begin{tabular}{c|ccc} 
\hline \hline
             & Optimal & Naive Partition & Heuristic Partition  \\ 
\hline
NASNet-A     & 3943    & 4013 ($\uparrow 1.78\%$)    & \textbf{3943}        \\
AmoebaNet-A  & 7732    & 7962 ($\uparrow 2.98\%$)    & \textbf{7732}                \\
Darts        & 1764    & \textbf{1764}               & \textbf{1764}         \\
HRNet-W18-V1 & 6368    & 7168 ($\uparrow 12.6\%$)    & \textbf{6368}                 \\
HRNet-W18-V2 & 6396    & 6900 ($\uparrow 7.88\%$)    & 6432($\uparrow 0.56\%$)         \\
HRNet-W32    & 6368    & 7040 ($\uparrow 10.60\%$)   & \textbf{6368}               \\
\hline \hline
\end{tabular}
\end{table}

Table \ref{table:partition num} shows the impact of partitioning quantities on
the effectiveness and efficiency of the scheduling. Experiments shows that our
partitioning method can achieve optimal result on NASNet. While sub-optimal
scheduling results can occur in HRNet, these instances typically display only
minor deviations from the optimal results. Additionally, it is worth noting
that even the sub-optimal peak memory footprint is significantly lower
compared to HMCOS, which achieves 7936 KB.

\begin{table}[!tb]
\centering
\caption{Peak memory footprint and scheduling time under different partitioning quantities}
\label{table:partition num}
\begin{tabular}{c|cc|cc} 
\hline \hline
\multicolumn{1}{c|}{} & \multicolumn{2}{c|}{HRNet-W18-V1}  & \multicolumn{2}{c}{NASNet-A}  \\
Part                 & Mem(KB)                 & Time(s) & Mem(KB)       & Time(s)       \\ 
\hline
1                    & \textbf{6368}           & /       & \textbf{3943} & 155           \\
2                    & \textbf{6368}           & 2832    & \textbf{3943} & 41.6          \\
3                    & \textbf{6368}           & 64      & \textbf{3943} & 61.7          \\
4                    & \textbf{6368}           & 1.12      & \textbf{3943} & 39.4          \\
5                    & \textbf{6368}           & 0.63    & \textbf{3943}        & 31         \\
6                    & 6400 ($\uparrow 0.5\%$)           & 0.17    & \textbf{3943} & 5.01          \\
\hline \hline
\end{tabular}
\end{table}

\section{RELATED WORK}

\textbf{Memory-aware scheduling.}
  Memory-aware scheduling can be classified into graph level and operator level. In the graph level, early work propose scheduling methods for trees or series-parallel graphs\cite{liu1987application,kayaaslan2018scheduling}.  SERENITY\cite{ahn2020ordering} first leverages dynamic programming in general network scheduling for peak memory footprint minimization. HMCOS\cite{wang2022hierarchical} proposes strategies for the hierarchization of a computation graph by detecting the NASNet-cells manually. In the operator level, FlexTensor\cite{zheng2020flextensor} optimize the tensor computation on heterogeneous systems.

\textbf{Neural network partition.}
Existing researches on neural network partition mainly focus on communication latency reduction for distributed deep learning\cite{mayer2017tensorflow, mayer2016graph,zheng2023soc}. ParDNN\cite{2020A} focuses on placing DNN’s underlying computational graph operations across multiple devices to satisfy the memory constraints. Our method is the first method using graph partition in memory-aware network scheduling. 

\textbf{Neural network optimization.}
Many optimizations make contributions to simplifying the DNN, such as parameter pruning\cite{han2015learning},  quantization\cite{jacob2018quantization,zhuang2020training}, low-rank Factorization\cite{kim2015compression}, and so on\cite{cai2022enable,zheng2023chimera}. The memory-aware scheduling research is orthogonal to these works.

\section{CONCLUSION}
This paper presents a memory-aware scheduling method for complex wired networks
on memory-constrained edge devices. This method employs an iterative 
sub-graph fusion technique to simplify the computation graph while preserving
scheduling optimality. A heuristic graph partitioning approach is also proposed
to further reduce the graph size when more efficient scheduling is needed.
The experimentation demonstrates
that our method can effectively and efficiently reduce the peak memory
footprint for complex neural networks.

\bibliographystyle{IEEEtran}
\newpage
\bibliography{reference/schedule.bib}

\end{document}